\documentclass{article}

    \PassOptionsToPackage{numbers, compress}{natbib}

\usepackage[dblblindworkshop, final]{neurips_2025}

\usepackage[utf8]{inputenc} 
\usepackage[T1]{fontenc}    
\usepackage{hyperref}       
\usepackage{url}            
\usepackage{booktabs}       
\usepackage{amsfonts}       
\usepackage{nicefrac}       
\usepackage{microtype}      
\usepackage{xcolor}         
\usepackage{kotex}
\usepackage{enumitem}
\usepackage{float}
\usepackage{graphicx}
\usepackage{array}
\usepackage{amsmath}
\newcommand{\codefrag}[1]{\mbox{\texttt{\detokenize{#1}}}\allowbreak}
\newcommand{\litn}{%
  \unskip
  \mbox{\texttt{\detokenize{\n}}}\nobreak
  \ignorespaces
}
\usepackage{float}
\workshoptitle{Efficient Reasoning}

\title{Studying the Korean Word-Chain Game with RLVR: Mitigating Reward Conflicts via Curriculum Learning}

\author{%
  Donghwan Rho \\
  Department of Mathematical Science\\
  Seoul National University\\
  \texttt{donghwan\_rho@snu.ac.kr} \\
}

\begin{document}

\maketitle

\begin{abstract}

Reinforcement learning with verifiable rewards (RLVR) is a promising approach for training large language models (LLMs) with stronger reasoning abilities. It has also been applied to a variety of logic puzzles. In this work, we study the Korean word-chain game using RLVR. We show that rule-derived rewards can naturally conflict, and demonstrate through experiments that a curriculum-learning scheme mitigates these conflicts.  Our findings motivate further studies of puzzle tasks in diverse languages.

\end{abstract}

\section{Introduction}

Recently, RLVR \cite{lambert2024tulu} has emerged as a promising approach to enhance the reasoning capabilities of LLMs, particularly in mathematics and coding \cite{guo2025deepseek,qwen2025qwen25technicalreport,yang2025qwen3}. Moreover, recent work \cite{chen2025enigmata,stojanovski2025reasoning,xie2025logic} has applied RLVR to several puzzle domains, offering solutions to tasks such as Sudoku \cite{chen2025enigmata} and Knights \& Knaves (K\&K) \cite{xie2025logic}. However, most of these studies concentrate on logical puzzles rather than language-centric ones. Even for language-related puzzles such as crosswords \cite{chen2025enigmata}, the focus is predominantly English. Consequently, puzzles in non-English or low-resource languages have received limited attention.

In parallel with algorithmic advances, the need for efficient training methods has grown, because training and evaluating LLMs incur substantial computational and memory costs. One line of work explores curriculum learning \cite{freitag2024curriculum}. For example, curricula can reorder training instances by problem difficulty \cite{xie2025logic} or gradually increase the distance between the start and the goal states \cite{florensa2017reverse}. These approaches have been shown to effectively improve the performance of LLMs.

In this work, we study the Korean word-chain game. Through this setting, we show that rewards derived from the rules of the game can conflict with one another, and that a curriculum-learning scheme mitigates these conflicts. Our study highlights the necessity for reward conflicts and that puzzles in non-English languages offer fertile directions for future research.

\subsection{Related work}

\paragraph{Reinforcement learning with verifiable rewards (RLVR).}
RLVR \cite{lambert2024tulu} has been used extensively to train models for reasoning tasks such as mathematics and coding \cite{guo2025deepseek,yang2025qwen3}. RLVR dispenses with a learned reward model in reinforcement learning (RL) \cite{sutton1998reinforcement} and instead uses verifiable rewards, typically binary (1 if correct; 0 otherwise). While RLVR is commonly applied to tasks with well-defined ground truth (e.g., mathematics), it can also be used for open-ended tasks \cite{hossain2024reinforcement}. In this work, we study the Korean word-chain game, which is rule-based yet open-ended.

\paragraph{Curriculum learning and reward conflicts in RL.}
Several works combine curriculum learning \cite{soviany2022curriculum} with RL \cite{freitag2024curriculum,jiang2025vcrl}. When multiple rewards are present in an RL task, they may conflict. For example, \cite{roijers2013survey} surveys sequential decision making under multiple and possibly conflicting objectives. Closest to our setting, \cite{freitag2024curriculum} introduces curriculum learning to reduce reward conflicts in RL tasks. Whereas their approach learns simpler rewards first to accelerate the discovery of successful trajectories, our approach learns a harder reward first to mitigate conflicts and acquire the rules more effectively.

\subsection{Contribution}

We investigate the Korean word-chain game using RLVR and make the following contributions:
(i) we show that rule-induced rewards can conflict intrinsically; and
(ii) demonstrate that, while naively applying RLVR with the full rule set fails to train a model effectively, a curriculum learning improves training efficiency by mitigating reward conflicts.

\section{Preliminary: Korean word-chain}\label{sec:korean_word_chain}

We study the Korean word-chain game, a language-specific variant of the English word-chain game. In english, the rule is simple: just propose a word with starts with the last character of the given word (e.g., `change'$\rightarrow$`energy'). In Korean, similarly we propose a Korean word which starts with the last syllable of the previous word; e.g., ``인사'' (greeting)$\rightarrow$``사랑'' (love). Unlike the English variant, the Korean game is subject to additional rules:

\begin{enumerate}[label=(\roman*)]
    \item The proposed word has to start with the first syllable of the last of the previous word.
    \item We can apply the \textit{initial-sound rule} which is called `두음법칙' in Korean.
    \item We must propose a noun from the Standard Korean Dictionary \cite{korea_dictionary} (표준국어대사전; hereafter, the \textit{dictionary}).
    \item Previous used words cannot be repeated.
\end{enumerate}

The initial-sound rule in Korean phonology refers to a convention where the consonants `ㄴ' and 'ㄹ' are avoided in certain conditions. For example, a word ``녀자''(woman) is realized as ``여자''. For the detailed explanation for the initial-sound rule, refer to Appendix \ref{sec:initial_sound_rule}. Hereafter, we simply refer to this task as \textit{word-chain} unless otherwise specified.

\section{Word-chain as a multiple rewards RLVR and rule conflict}\label{sec:word_chain_multiple_rewards}

Word-chain is a rule-based yet open-ended task: many answers are possible, and each answer can be validated with verifiable rewards. While one can consider word-chain as a simple game of saying words, the model has to follow multiple rules to continue the game. In particular, the rules (i)--(iv) in Section \ref{sec:korean_word_chain} naturally induces the multiple rewards. Naively, one can set rewards as follows, which we call \textit{Baseline}:
\begin{enumerate}[label=(\alph*)]
    \item If the first syllable of the proposed word is the same as the last one of the previous word, add 1.
    \item If (a) is satisfied and the proposed word is a noun, then add 1.
    \item If the proposed word repeats any previously used word, the reward becomes -1.
\end{enumerate}

\begin{table}[H]
\small
\begin{center}

\caption{\label{tab:initial_sound_rule_failure_cases}The failure cases for the initial-sound rule. If we train the model with the naive rewards, the model fails to acquire the initial-sound rule.
}

\begin{tabular}{ccc}
\toprule
          Previous Word & Answer (Correctness) & After the initial-sound Rule (Correctness) \\ \cmidrule(lr){1-3}
          가동\textcolor{red}{력} & \textcolor{red}{력}량 (False) & \textcolor{blue}{역}량 (True) \\
          아\textcolor{red}{롱} & \textcolor{red}{롱}소 (False) & \textcolor{blue}{농}소 (True) \\
          고\textcolor{red}{료} & \textcolor{red}{료}액 (False) & \textcolor{blue}{요}액 (True) \\
          물리\textcolor{red}{량} & \textcolor{red}{량}수 (False) & \textcolor{blue}{양}수 (True) \\
          
\bottomrule
\end{tabular}
\end{center}
\end{table}

Since the initial-sound rule is optional, we did not include the reward for it. However, we observe the model fails to acquire this rule. We guess that this is because the first rule is much easier than the second rule. Table \ref{tab:initial_sound_rule_failure_cases} illustrates such cases; the model outputs words to which the initial-sound rule is not applied. They are failure cases because they are not nouns. If the model applies the initial-sound rule to the outputs, the answers are valid. A fundamental source of reward conflict of the word-chain is as follows:

\[
\textit{The basic rule (i) and the initial-sound rule (ii) conflict intrinsically.}
\]

This is not resolved easily just by training a model. In training, we observe that for the inputs for which we can apply the initial-sound rule, the answers only follow the first rule, therefore resulting in the same rewards, making no signal for the initial-sound rule. In this way, the reward for (a) interferes learning the initial-sound rule. We observe the same phenomenon even if we change the scales of the rewards.

To mitigate this, we enforce the initial-sound rule in the rewards, which we term \textit{initial-sound Rule Forcing}:

\begin{enumerate}[label=(\alph*)]
    \item[(a-i)] If the first syllable $\mathcal{F}$ of the proposed word is the same as the last syllable $\mathcal{L}$ of the previous word, add 1.
    \item[(a-ii)] If we can apply the initial-sound rule to $\mathcal{L}$ and is the same as $\mathcal{F}$, add 1.
    \item[(b)] If (a-i) or (a-ii) holds and the proposed word is a noun, then add 1.
    \item[(c)] If the proposed word repeats any previously used word, the reward becomes -1.
\end{enumerate}

In other words, if the initial-sound rule can be applied, this rule must be applied for the full reward. This reward variation improves the word-chain capability of the model and it will be explained in Section \ref{sec:experimental_results}.

\section{Preventing rule conflict via data reordering}\label{sec:data_reordering}

\begin{figure}[H]
\begin{centering}
\includegraphics[width=0.99\textwidth]{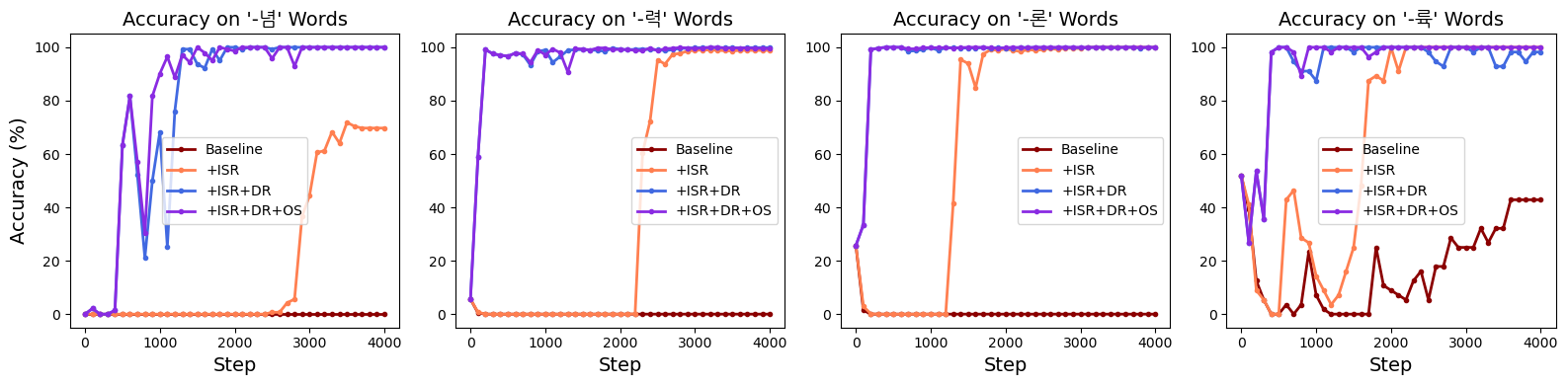}
\par\end{centering}
\caption{\label{fig:initial_sound_rule_case_test}
Accuracy by final-syllable category. For example, the leftmost graph shows the accuracy of model answers for words which ends with ``념''. In this figure, `념', `력', `론', `륙' are syllables to which we can apply the initial-sound rule. While the baseline fails to learn or learns the initial-sound rule too slowly, our methods ISR, DR, and OS (see Section \ref{sec:results_analysis}) accelerate acquiring the rule.
}
\end{figure}

Although enforcing the initial-sound rule improves the word-chain ability of the model, the model still struggles to acquire this rule; for each syllable, it fails to acquire the rule, or acquires too slowly (see Figure \ref{fig:initial_sound_rule_case_test}).
Too address this, we adopt a two-stage curriculum: at the first stage, the model is trained only on examples where the initial-sound rule applies, promoting learning the initial-sound rule in the early stage of training. Next, in the second stage, the model is trained with the all examples. At the second stage, the first and the second rewards does not conflict because model already acquired the initial-sound rule at the first stage. In addition, because acquiring the initial-sound rule is harder than other rules, we increase the proportion of the initial-sound rule examples in training.

\section{Experimental Results}\label{sec:experimental_results}

We train using RLVR under the different rewards settings described in Sections \ref{sec:word_chain_multiple_rewards} and \ref{sec:data_reordering}. We use \texttt{beomi/Qwen2.5-7B-Instruct-kowiki-qa} \cite{beomi_qwen25_kowiki_qa}, an instruction fine-tuned version of \texttt{Qwen2.5-7B} \cite{qwen2025qwen25technicalreport} with Korean QA dataset such as \cite{ko_wikidata_QA} as our base model because the task is in Korean. We train for 4000 steps.

\subsection{Dataset generation}

\begin{table}[H]
\small
\centering
\caption{\label{tab:format}The input format used in training.}
\begin{tabular}{@{}p{.48\textwidth} p{.48\textwidth}@{}}
\toprule

\multicolumn{2}{c}{Input Format} \\ \midrule
\multicolumn{2}{@{}p{.96\textwidth}@{}}{%
  \raggedright\ttfamily
  \codefrag{<|im_start|>system}\litn%
  \codefrag{{Prompt}<|im_end|>}\litn
  \codefrag{<|im_start|>user}\litn
  \codefrag{{Word 1}<|im_end|>}\litn
  \codefrag{<|im_start|>assistant}\litn
  \codefrag{{Word 2}<|im_end|>}\litn
  \,\(\cdots\)\, 
  \codefrag{<|im_start|>user}\litn
  \texttt{\{}\texttt{Word }\(2N-1\)\texttt{\}}\allowbreak
  \codefrag{<|im_end|>}\litn
  \codefrag{<|im_start|>assistant}%
} \\
\bottomrule
\end{tabular}
\end{table}

For training and evaluation, we provide a prompt specifying the rules (see Table \ref{tab:prompt} in Appendix \ref{sec:dataset_generation}). The input format is provided in Table \ref{tab:format}.With the prompt and the format, we make a train dataset. Each example contains a chain with $2N-1$ words for a randomly selected $N$. We add special tokens to train the model in the multi-turn conversation setting. The model predicts the $2N$-th word. For detailed explanation for dataset generation, refer to Appendix \ref{sec:dataset_generation}.

\subsection{Observation: starting with non-final syllable}\label{sec:other_syl}

\begin{table}[H]
\small
\begin{center}

\caption{Failure case when an answer starts with the syllable of a previous word, but not the last one. If the epoch is low, this case arises frequently.\label{tab:other_syllable}
}

\begin{tabular}{cc}
\toprule
          Previous Word & Answer \\ \cmidrule(lr){1-2}
          식\textcolor{red}{문}화 & \textcolor{red}{문}화재 \\
          \textcolor{red}{산}업 & \textcolor{red}{산}업화 \\
          트램\textcolor{red}{웨}이 & \textcolor{red}{웨}이트 \\
          출판\textcolor{red}{문}화 & \textcolor{red}{문}화유산 \\
          
\bottomrule
\end{tabular}
\end{center}
\end{table}

In training and evaluation, we observe that when the model fails to answer, it often outputs a word which starts with a syllable of the previous word that is different from the last one. (See Table \ref{tab:other_syllable}). This failure cases arise with nontrivial frequency as shown in Figure \ref{fig:failure_cases}. To mitigate this behavior, wee add another reward (d) and term the setting including rewards (a-i)--(d) \textit{Other Syllable}:
\begin{enumerate}
    \item[(d)] If the answer starts with a syllable of the previous word that is not the last one, add -0.5.
\end{enumerate}

\subsection{Word-chain game versus the dictionary}\label{sec:vs_dictionary}

\begin{figure}
\begin{centering}
\includegraphics[width=0.95\textwidth]{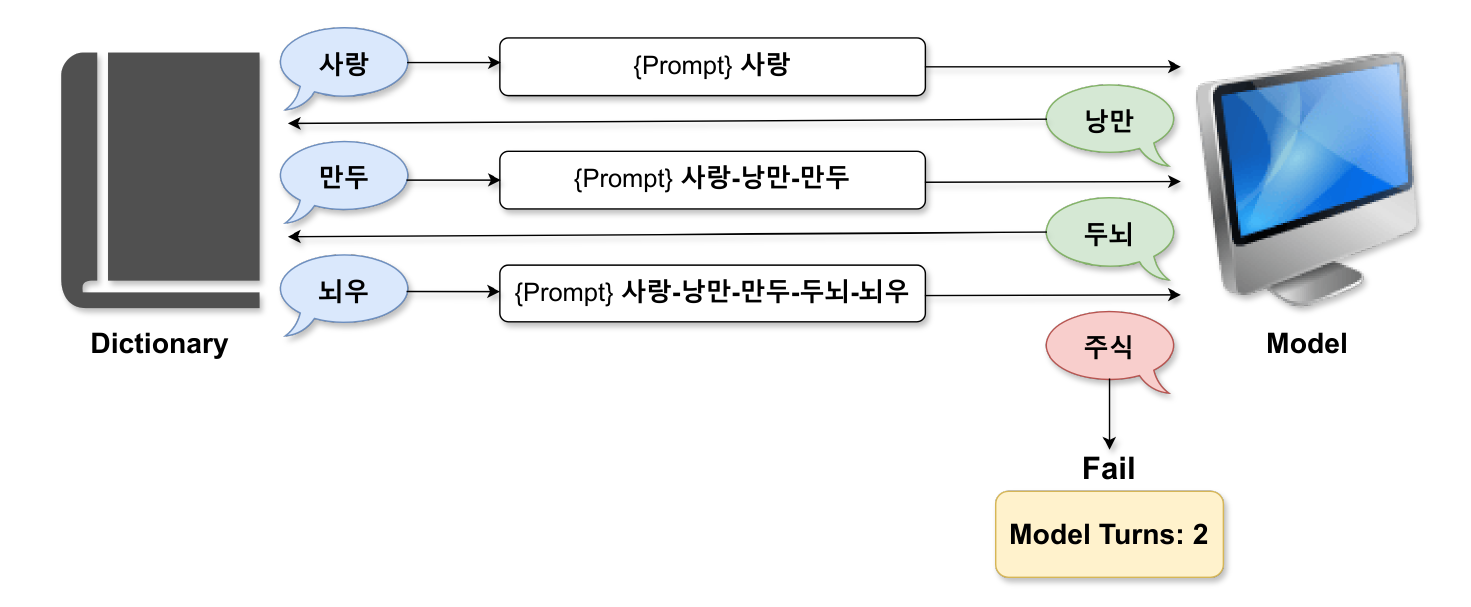}
\par\end{centering}
\caption{\label{fig:word_chain_game}
A word-chain game against the dictionary. The dictionary starts with a random word (e.g., ``사랑''). Then this word is sent to the model together with the prompt. The model predicts the next word (``낭만'', applying the initial-sound rule). If the next word satisfies the rules, the dictionary chooses another noun and the game continues until either the dictionary cannot supply a valid word or the model violates a rule. In this game, the model fails to answer the third word, and the model turn is 2. \textit{The meaning of Korean words}: ``사랑'': love, ``낭만'': romance, ``만두'': dumpling, ``두뇌'': brain, ``뇌우'': thunderstorm, ``주식'': stock.
}
\end{figure}

To evaluate, we play the word-chain game against the dictionary introduced in Section \ref{sec:korean_word_chain}.  The process of the game is described in Figure \ref{fig:word_chain_game}.

\subsection{Results analysis}\label{sec:results_analysis}

\begin{figure}[H]
\begin{centering}
\includegraphics[width=0.95\textwidth]{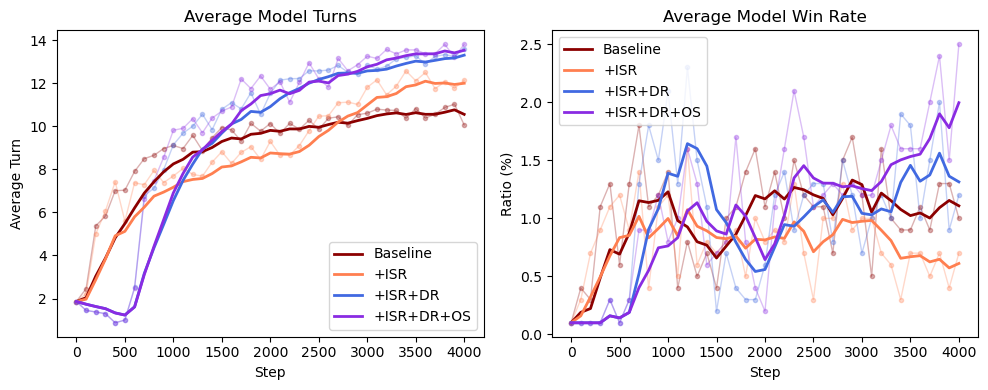}
\par\end{centering}
\caption{\label{fig:results_no_repetition}
\textbf{(Left)} the averaged model turns and \textbf{(Right)} the average winning rate of the model in the word-chain game versus the dictionary. According to the proposed methods, both of metrics increase.
}
\end{figure}

We analyze performance using two metrics: average model turns and win rate. For each checkpoint and setting, we run 1000 games against the dictionary. For example, if a model answered the $n$-th word correctly but failed at the $(n+1)$-th turn, then the turn count is $n$. If the dictionary cannot choose a next word, the model wins. When the model wins, we do not include this game in computing metrics.

As stated in the previous sections, we consider four additive settings: \textit{Baseline} (Section \ref{sec:word_chain_multiple_rewards}), \textit{initial-sound Rule Forcing} (ISR, Section \ref{sec:word_chain_multiple_rewards}), \textit{Data Reordering} (DR, Section \ref{sec:data_reordering}), \textit{Other Syllable} (OS, Section \ref{sec:other_syl}). 
Through experiments, we have the following results as in Figure \ref{fig:results_no_repetition}:
\begin{enumerate}[label=(\roman*)]
    \item Under Baseline setting, the average turns does not increase slowly. In contrast, as we add our methods (ISR, DR, OS), the performance improves modestly; we can see that OS also increases the average turns slightly.
    \item Win rate is noisy, but when all of our methods are combined, the final win rate is highest. However, ISR alone does not seem to increase win rate.
    \item The model exhibits \textit{length-generalization}: although each training example consists of at most seven words (that is, four turns; see Appendix \ref{sec:dataset_generation}), average turns significantly exceed four.
\end{enumerate}

\begin{figure}[H]
\begin{centering}
\includegraphics[width=0.99\textwidth]{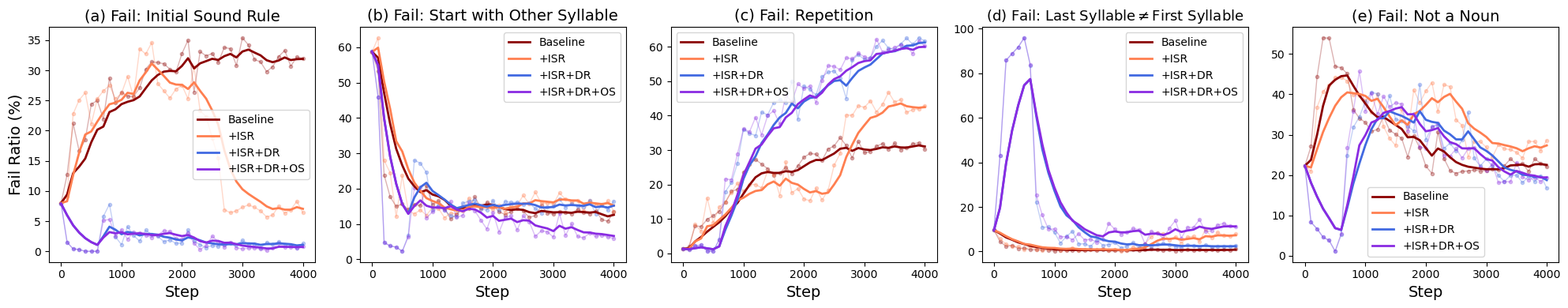}
\par\end{centering}
\caption{\label{fig:failure_cases}
Failure cases in the word-chain game against the dictionary. \textbf{(a)} the answer does not obey the initial-sound rule; \textbf{(b)} the answer starts with the syllable of the previous word that is different from the last one (Section \ref{sec:other_syl}); \textbf{(c)} the answer is one of the previous words; \textbf{(d)} the answer does not follow the rule (i) in Section \ref{sec:korean_word_chain}; \textbf{(e)} the answer is not a noun.
}
\end{figure}

Moreover, we analyze the failure modes described in Figure \ref{fig:failure_cases}.
Given an answer, multiple failure modes can arise. For precise analysis, we prioritize cases, as specified in Appendix \ref{sec:failure_cases_order}. If the model wins a game, then a failure count does not increase; for example, if the model wins 50 games among 1000 runs, then the sum of all of the failure mode ratios is 0.95. Our results for Figure \ref{fig:failure_cases} are as follows:
\begin{enumerate}[label=(\alph*)]
    \item Under Baseline setting, the model struggles to learn the initial-sound rule. Even under ISR setting, the failure ratio for the initial-sound rule (a) does not decrease until around 2000 epoch. This decrease after 2000 epoch accords to the increasing tendency of average model turns in ISR setting after 2000 epoch (see Figure \ref{fig:results_no_repetition}). However, with DR, the model quickly learns this rule.
    \item One may think that the reward for other syllable in Section \ref{sec:other_syl} is redundant because if the model successfully learns the rule (i) in Section \ref{sec:korean_word_chain}, the failure ratio for (b) should automatically decrease. However, we observe that in other settings without OS, the failure ratio for (b) is larger than 0.1. and with OS, this ratio decreases about half. In contrast, at the same time, the failure ratio for (d) slightly increases, while we do not establish a causal link between them.
    \item It seems that the model fails to learn the repetition rule as shown in Figure \ref{fig:failure_cases}. We guess this is because the model hardly outputs a repeated answer during training. We leave acquiring the repetition rule as a future work.
    \item Since the model in only trained with the initial-sound rule examples under the first stage of DR setting, it only outputs words related to the initial-sound rule at the early stage of training, increasing the failure ratio for (d). However, this sharply decreases as the training proceeds on the full dataset.
\end{enumerate}

\section{Conclusion}

This work studies the Korean word-chain game and the reward conflicts naturally arising from the rules of the game. To address this problem, we propose a curriculum to learn a harder rule first. Experimental results demonstrate that our methods can significantly improve the model performance and reduce reward conflicts. We believe that studying non-English puzzle tasks can motivate a lot of interesting ways to enhance the reasoning capabilities of LLMs.

\bibliography{bib}

\appendix

\section{The initial-sound rule}\label{sec:initial_sound_rule}

The initial-sound rule isa phonological convention whereby certain syllables can be avoided when they are the first syllables of a word. The initial consonants for the initial-sound rule is `ㄴ, ㄹ' and they can be changed to 'ㄴ, ㅇ'. However, all of the syllables which start with these consonants cannot be changed. All of the cases of the initial-sound rule is in Table \ref{tab:initial_sound_rule}. The initial-sound rule can be applied when a final consonant exists. For example, ``리치'' (reason) and ``림업'' (forestry) can be changed to ``이치'' and ``임업'', respectively.

\begin{table}[H]
\small
\begin{center}

\caption{\label{tab:initial_sound_rule}Cases for the initial-sound rule.
}

\begin{tabular}{cc}
\toprule
          Before & After \\ \cmidrule(lr){1-2}
          녀 & 여 \\
          뇨 & 요 \\
          뉴 & 유 \\
          니 & 이 \\
          라 & 나 \\
          래 & 내 \\
          랴 & 야 \\
          려 & 여 \\
          례 & 예 \\
          로 & 노 \\
          뢰 & 뇌 \\
          료 & 요 \\
          루 & 누 \\
          류 & 유 \\
          르 & 느 \\
          리 & 이 \\
\bottomrule
\end{tabular}
\end{center}
\end{table}

Originally, the initial-sound rule is stated for words. However, in this work, we apply this rule as a rule that a certain syllable can be changed to as another syllable.

\section{Experimental details}\label{sec:experimental_details}

In this work, we train the model with LoRA \cite{hu2022lora}. We describe the train hyperparameters in Table \ref{tab:model_hyperparameters}.

\begin{table}[H]
\small
\begin{center}

\caption{\label{tab:model_hyperparameters}The model and hyperparameters used in the training.
}

\begin{tabular}{cc}
\toprule
          Model & \texttt{beomi/Qwen2.5-7B-Instruct-kowiki-qa} \cite{beomi_qwen25_kowiki_qa} \\ \cmidrule(lr){1-2}
          LoRA Rank & 16 \\ \cmidrule(lr){1-2}
          \multicolumn{2}{c}{Generation Configuration} \\ \cmidrule(lr){1-2}
          \texttt{temperature} & 1.0 \\
          \texttt{top\_$p$} & 1.0 \\
          \texttt{max\_new\_tokens} & 20 \\ \cmidrule(lr){1-2}
          \multicolumn{2}{c}{Optimizer} \\ \cmidrule(lr){1-2}
          Optimizer & \texttt{AdamW} \\
          Learning Rate & $1\times10^{-4}$ \\
          $\beta_1$ & 0.9 \\
          $\beta_2$ & 0.99 \\
          Weight Decay & 0.1 \\
          Warmup Ratio & 0.1 \\
          LR Scheduler Type & \texttt{cosine} \\ \cmidrule(lr){1-2}
          \multicolumn{2}{c}{GRPO Configuration} \\ \cmidrule(lr){1-2}
          \texttt{max\_prompt\_length} & 256 \\
          \texttt{max\_completion\_length} & 20 \\
          \texttt{max\_grad\_norm} & 5 \\
          Number of Generations Per Prompt & 4 \\
          Batch Per Device & 1 \\
          Gradient Accumulation Step & 32 \\
\bottomrule
\end{tabular}
\end{center}
\end{table}

\section{Dataset generation}\label{sec:dataset_generation}

We describe the dataset generation process. First, we state the prompt used in training and evaluation in Table \ref{tab:prompt}. We combine the prompt with a chain of words. A chain consists of words following the rules of the word-chain, and the length of each chain is in $\{1,3,5,7\}$ in this work. Because the initial rule is harder than the basic rule (i), we select the general nouns and nouns to which the initial-sound rule is applicable in different ratios. More precisely, we classify nouns according to the last syllable. After that, from each group $G$ of words which end with a syllable $\mathcal{L}$, we choose a $k$ elements, where
\begin{align*}
    k=|G| & \quad \text{ if $|G|<10$,} \\
    k=\min(\max(\lfloor 0.01\times|G| \rfloor, 10),200) & \quad \text{ if $|G|\ge 10$ and $\mathcal{L}$ is a general syllable, or} \\
    k=\min(\max(\lfloor 0.2\times|G| \rfloor, 10),200) & \quad \text{ if $|G|\ge 10$ and we can apply the initial-sound rule to $\mathcal{L}$,}
\end{align*}
where $\lfloor\cdot\rfloor$ means truncation toward zero. In this way, we can choose words with appropriate numbers. In total, we use only about 5.2\% of the nouns in the dictionary for training.

\begin{table}[H]
\small
\centering
\caption{\label{tab:prompt}Korean/English prompts used for training and evaluation of word-chain. We used the Korean prompt in training and evaluation and the English prompt is provided for understanding.}
\begin{tabular}{@{}p{.48\textwidth} p{.48\textwidth}@{}}
\toprule
\multicolumn{1}{>{\centering\arraybackslash}p{.47\textwidth}}{\textbf{Korean Prompt}} &
\multicolumn{1}{>{\centering\arraybackslash}p{.47\textwidth}}{\textbf{English Prompt}} \\
\midrule
당신은 친절한 영어 인공지능 비서입니다. 우리는 단어 끝말잇기를 할 것입니다. 제가 단어를 말하면, 그 단어의 마지막 글자로 시작하는 두 글자 이상의 한글 단어를 말해주세요. 이미 사용된 단어는 반복하지 마세요. 표준국어대사전에 있는 단어를 말해주세요. 두음법칙은 적용 가능하되, 적용 후의 단어를 말해주세요. 예를 들어 ‘력량’이 아닌 ‘역량’, ‘료리’가 아닌 ‘요리’라고 대답해야 합니다. 다른 문장은 쓰지 말고, 당신의 응답(단어)만 적어주세요. 
&
You are a helpful English AI assistant. We will play a Korean word-chain game. When I give a word, respond with a Korean word of two or more syllables that starts with the last syllable of my word. Do not repeat any words that have already been used. Use words that appear in the Standard Korean Language Dictionary. The initial-sound rule may apply; if it does, respond with the form \emph{after} applying it—for example, answer ``역량” instead of “력량”, and “요리” instead of “료리”. Do not write any other sentences; output only your response (the word). \\
\bottomrule
\end{tabular}
\end{table}

\section{Priority order for failure modes}\label{sec:failure_cases_order}

\begin{figure}[H]
\begin{centering}
\includegraphics[width=0.99\textwidth]{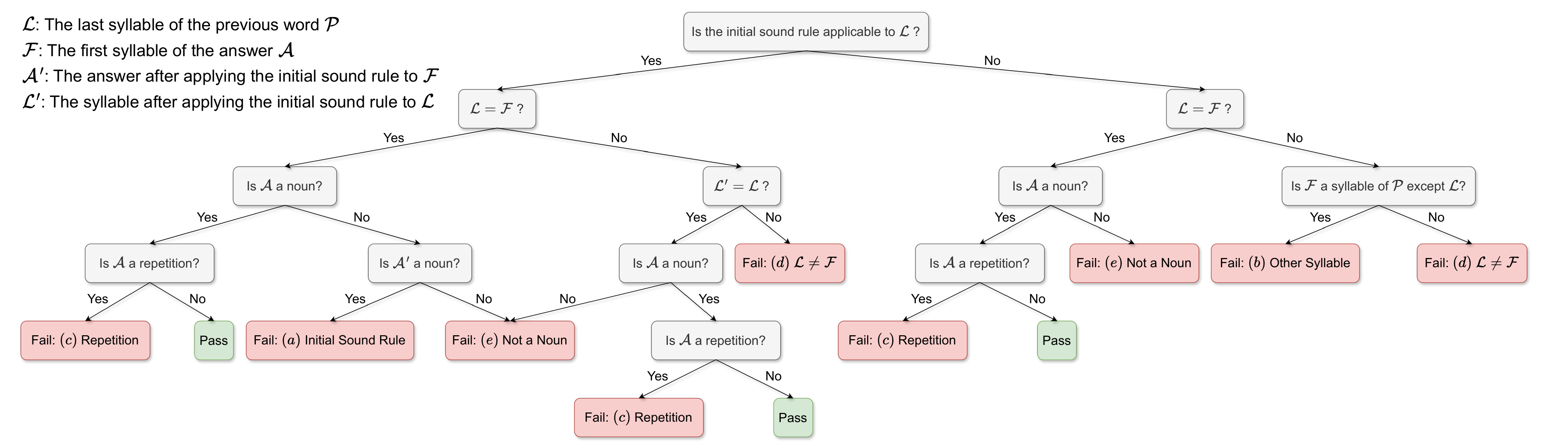}
\par\end{centering}
\caption{\label{fig:failure_priority}
The flow chart representing the classification of the failure modes.
}
\end{figure}

To prevent overlap of the failure modes in the word-chain game versus the dictionary in Section \ref{sec:vs_dictionary}, we set the priority. We provide the priority in Figure \ref{fig:failure_priority}.


\end{document}